\documentclass{article}
  \usepackage[a4paper]{geometry}
  \usepackage{cite}
  \usepackage[pdftex]{graphicx}
  \usepackage{amsfonts}
 
  \usepackage{times}
  \usepackage{url}
  \usepackage{array}
  \usepackage[font=footnotesize]{subfig}
  \usepackage{graphicx}
  \usepackage{mathtools}
  \usepackage{enumitem}

  \usepackage{alltt}

  \def\boxx{{\vcenter{\vbox{\hrule height.3pt
            \hbox{\vrule width.3pt height6pt
            \kern6pt\vrule width.3pt}\hrule height.3pt}}\;}}
 
  \def\impos{{\;\vcenter{\hbox{\rule{5mm}{0.2mm}}} \vcenter{\hbox{\rule{1.5mm}{1.5mm}}} \;}}

  \def\lrarrow{\leftrightarrow \kern-8pt \rightarrow}

  \def\2{\frac{1}{2}}
 
  \def\beq{\begin{eqnarray}}
  \def\eeq{\end{eqnarray}}
  \def\2{\frac{1}{2}}

  \newtheorem{definition}{Definition}

  \usepackage{CJKutf8}
  
\begin{document}
 
  \newcommand{\strust}[1]{\stackrel{\tau:#1}{\longrightarrow}}
  \newcommand{\trust}[1]{\stackrel{#1}{{\rm\bf ~Trusts~}}}
  \newcommand{\promise}[1]{\xrightarrow{#1}}
  \newcommand{\revpromise}[1]{\xleftarrow{#1} }
  \newcommand{\assoc}[1]{{\xrightharpoondown{#1}} }
  \newcommand{\imposition}[1]{\stackrel{#1}{\impos}}
  \newcommand{\scopepromise}[2]{\xrightarrow[#2]{#1}}
  \newcommand{\handshake}[1]{\xleftrightarrow{#1} \kern-8pt \xrightarrow{} }
  \newcommand{\cpromise}[1]{\stackrel{#1}{\frightarrow}}
  \newcommand{\policy}{\stackrel{P}{\equiv}}
  \newcommand{\field}[1]{\mathbf{#1}}
  \def\R1{\mathbb{R}}
  \newcommand{\bundle}[1]{\stackrel{#1}{\Longrightarrow}}

  \title{On The Role of Intentionality in Knowledge Representation\\~\\{\Large Analyzing Scene Context for Cognitive Agents\\ with a Tiny Language Model}}
 
  \author{Mark Burgess}
 \date{July 13, 2025}
  \maketitle
 
  \begin{abstract}
    Since Searle's work deconstructing intent and intentionality in
    the realm of philosophy, the practical meaning of intent has
    received little attention in science and technology.
    Intentionality and context are both central to the scope of
    Promise Theory's model of Semantic Spacetime, used as an effective
    Tiny Language Model. One can identify themes and concepts from a
    text, on a low level (without knowledge of the specific language)
    by using process coherence as a guide.  Any agent process can
    assess superficially a degree of latent `intentionality' in data
    by looking for anomalous multi-scale anomalies and assessing the
    work done to form them. Scale separation can be used to sort parts
    into `intended' content and `ambient context', using the spacetime
    coherence as a measure.  This offers an elementary but pragmatic
    interpretation of latent intentionality for very low computational
    cost, and without reference to extensive training or reasoning
    capabilities.  The process is well within the reach of basic
    organisms as it does not require large scale artificial
    probabilistic batch processing. The level of concept formation
    depends, however, on the memory capacity of the agent.
  \end{abstract}
 
 

  \section{Introduction} 
 
  Cognitive abilities, which include ideas like intentionality and
  consciousness, have long been viewed in Western philosophy as
  exclusive to the human realm. Intent is roundly considered
  justifiable only with minimum requirements for self-awareness or
  situational comprehension.  However, such hard line views have
  softened gradually with modern enlightenment, and more of us are likely
  to accept that terms such as `agency', `intelligence', and even
  `emotion' can apply for other species too. Even plants lean into
  sunlight in an intentional way; the identification of an intention
  doesn't have to arise from the plant to be true.  Latterly their
  possibility has been extended even to artificial systems, which some
  find more acceptable, though a modern version of the privilege
  argument persists in a distinction between `simple' machinery and
  `complex' biology, which many believe still holds some principled
  leap in understanding.  Ideological `blood-brain barriers', like
  these, continue to undermine efforts to form a rational causal
  explanation of intent, leading extremists to clutch at esoteric
  straws like quantum mechanics or complexity theory to account for
  perceived magic.

  In this note, I address another apparent schism that may shed light
  on these questions: the difference between process dynamics (the
  realm of physics) and interpretive semantics (the realm of
  linguistics and philosophy), and the suggestion that (deep down)
  intentionality might be a relatively simple phenomenon with an
  energetic explanation (as trust has been shown to
  be\cite{burgessdunbarpub}). The recent acceptance of attention mechanisms
in Large Language Models is related example\cite{attention0,attention1}.  On a large enough scale of action,
  where human thought tends to dwell, intent can be imagined as a
  vector pointing to a particular outcome in a virtual space of
  outcomes. This assumes that such a space could be constructed
  consistently, which is non trivial in practice. But what about on
  the scale of single sentence, or in reacting to a single image? What
  set of basis vectors would span such an idealized space? What
  palette of behaviours would it call into being?

For ideologues, the inability to
  capture precise ad hoc requirements, in every detail, means that a
  rational explanation can always be rejected---yet science's goal is
  to aim only for a suitably idealized approximation with predictive
  power.  This note addresses a pragmatic, elementary, and idealized
  definition of intent, motivated by a decade of studies based on 
  agent properties and Promise Theory\cite{promisebook}.  Promise Theory
  highlights the cooperative roles of dynamics and semantics in
  autonomous agents. Abstracting away human concerns, promises define
  how agents can state their intentions without the need for human
  language.  A door handle expresses its function through its form,
  for instance---which is intentional.

  The ability to detect `intentionality' is relevant in many areas
  today, not only Artificial Intelligence, but also in biology,
  neuroscience, sociology, even politics etc. A simple task like the
  indexing of a book relies on being able to predict the intent of a
  reader to identify or `resonate' with certain phrases that are
  intentional in the sense that they are introduced for a purpose.
  This is a crucial building block for knowledge representations and
  semantic graphs, which are poorly understood today.

  Here, I pursue the simplest dynamical definition of intentionality,
  based on process coherence and pattern repetition rates, and use
  this to distill a chemistry of semantic elements from `sensory
  input'. The outcome teaches us some counter-intuitive lessons. 
  By modern standards, this is a low-brow approach to cognition
  compared to the more usual brute force data of Large Language Models, Deep
  Learning, and behavioural emulation.  Nevertheless it can be used to
  cheaply, and without training, build robust knowledge graphs, which
  can then be queried using the same principles of the Semantic
  Spacetime model. It sidesteps awkward fixes like ontologies that
  require the designing and testing and have logical flaws, and builds on earlier work in
\cite{burgess2020testingquantitativespacetimehypothesis1,burgess2020testingquantitativespacetimehypothesis2}. 
For this analysis we need only a Tiny Language Model (TLM), which doesn't even know
  about the words of specific languages per se.

  After defining some assumptions and concepts, we apply the methods
  directly to Natural Language Processing based on some narrative
  texts (because these are readily available and are of direct
  practical use in the SSTorytime Project for knowledge
  representation\cite{sstorytime}).  I discuss the method of symbolic
  fractionation and interferometry for distinguishing ambient parts
  from anomalous parts of data. Finally, we end with some conclusions
  about the importance of dynamic over semantic reasoning.

\section{Definitions and assumptions}

Searle wrote the highly influential treatise on intentionality in
\cite{searle1}, which covers much ground, but never ventures into the
realm of process mechanics. Despite this, there exists a direct
dimensional analogy between promises, intent, and energy-momentum in
physics, which has been discussed previously (e.g. see
\cite{virtualmotion2}). Here, I will focus on the physics of information
rather than the epistemology.

Intent is associated with goals, i.e.
directed efforts towards an end in an abstract space of outcomes. It
has much in common with the ballistic directionality of momentum in
physics. Intentionality, on the other hand, refers to the general
capacity to project, induce, or adopt such intentions (more like the
accounting parameter of energy, potential summarizing a history of
kinetic).  When a scenario is charged with intentionality, it has the
potential to move an observer's process in a specific direction,
features we associated with purely causal phenomena like potential,
force, and momentum in physics\cite{trustnotes}.  The Semantic
Spacetime model\cite{spacetime1,spacetime2,spacetime3} was developed
as a bridge between basic causal thinking and higher level ideas
relating to semantics and cognition. In earlier work I showed how one
could test its hypotheses for Natural Language
Processing\cite{NLP,burgess2020testingquantitativespacetimehypothesis1,burgess2020testingquantitativespacetimehypothesis2}
to search for concepts and themes based on energy principles rather
than Markov models and training.

In a cognitive scenario, where intent plays a role, we are concerned
with sensing and representing some kind of scene involving other
agents, sampling the environment, receiving visual, auditory,
information etc. In all cases, an agent must be able to detect a
signal from outside itself whose behaviours one may or may not wish to
describe as intentional.  The assessment of intentionality comes from
examining the features of a scene or episode, experienced through some
sensory channel. We should immediately take care to distinguish from
the idea of `sentiment analysis', which is used in NLP to assess the
emotional tone of writing, based on a text analysis and a knowledge of
language and its accepted meanings.  Intentionality is not about
asking what the actors and subjects of a scene might be thinking or
feeling (if they can even think or feel); rather, it's about the
extent to which their intentional behaviour is distinguishable from
its ambient background in order to distinguish action from passive
background. Does the scene unfold ad hoc, or in a more purposeful
manner, and how can we tell the difference?

With this caveat, we begin with some clarifications of terminology.
Intent is associated with processes that are directed and aligned
towards a coherent outcome.  Random variables will have low
intentionality, and focused signalling may have high intentionality.
For example, think of the Search for Extra Terrestrial Intelligence
(SETI), scanning for signals in a noisy background. How would one
assess that a signal possessed of intent and was not merely a regular
pattern from a pulsar? Repetition may be part of intent (as a form of
emphasis, like speaking loudly), but duplication alone is not enough
to draw attention.

\begin{definition}[Intentionality]
Intentionality is a capacity to represent intent.
A phenomenon may be described as intentional if it has a measurable intentionality
regardless of the source.
\end{definition}
Agent observers, then, are the arbiters of
intentionality in a source, whereas source agents are the originators of the intent
being assessed.
We can refer to as a cognitive agent.
\begin{definition}[Cognitive agent]
Any attentive agent, receiving input from its ambient exterior, processing it, and
reacting to it.
\end{definition}
Cognition is the process of sensing, sampling, and responding to a
scene.  This universal interpretation recognizes `agency' simply as
the causal independence of behaviour (a situated form of what physics
would refer to as `locality'), which includes the capability of an agent
to engage independently in reactive behaviour after processing some
form of sensory information. This does not require any self-awareness
or elite club membership.  Even an RNA mechanism can deconstruct a string
of DNA and respond to it without its awareness or understanding. A
single atom absorbing photons, single cells, higher animals, up to
entire cities, can all be described as cognitive agents, with respect
to certain processes, identifiable as an entity by some appropriate
boundary.  Cognition, then, does not imply a lower bound on agent
`intelligence', nor a specific definition of `comprehension'.

\section{Heuristics for intent as thermodynamic work}

We can now sketch some heuristics for intent.  Intent concerns a
process of many small parts, all of which are possible to combine in a
fluid way to embellish the semantics of a process in the eyes of some
observer. It is never important what the causal originator of a
behaviour intends, we can only assess what an observer sees in that
behaviour.  This combinatoric state space is typical of a statistical
system. However, unlike traditional thermodynamics, an intentional
system is not a system that we can claim (in any sense) to be in
equilibrium. Intent is more like the effort to deviate from a
predictable balance, so probabilities will not suffice to describe it.

Intent always originates from an active agent source $S$:
\beq
S \promise{+p} R
\eeq
An receiving or observing agent $R$ assesses the
intentionality of a process $p$, originating from another agent, but it does
so using its own intentional processes, investing
effort to pay attention and decode the scene:
\beq
R \promise{-p} S.
\eeq
In Promise Theory, these are called complementary promises. Two agents'
intentions are needed to make such an assessment.
All agents are potentially intentional in their behaviours.
If behaviours are simple, the intent
may be trivial; if the processes are complicated, intent is more complicated.

As mentioned above, the concept of intentionality is something like a
potential energy surface.  Kinetic energy and momentum refer to
phenomena that are happening `now', whereas potential energy is a
representation of all prior history leading up the present
configuration of circumstances.  Physical potential representations
flatten all previous history into an ambient snapshot of what can
happen next. Intentionality is a relative property, like a potential,
that we use to measure the degree of non-randomness over some scale.
It has no absolute value: only differences matter.  Many physical
processes are of the Markov type, meaning that they are
memoryless---they forget past history and decouple.  Other, more
complex systems, retain memory of past exchanges internally and behave
accordingly. Agents (possessing interior resources) can represent both
kinds.

Let us postulate the existence of a potential function:
\beq
I(w,\ldots)
\eeq
which can be identified with the intentionality of an observation $w$
in a stream of such $w_i$. This might represent an underlying process of any kind.
How might we define the intentionality of $w$ as such a part of a stream?
\begin{itemize}
\item Singleton, spurious, or unique events that do not invest significant work cost
would have low intentionality. One might even consider them to be
errors, accidents, cosmic rays, etc\footnote{The question of errors
  and accidents is interesting. One could try to argue that errors are
  specifically unintended (a question of policy), but without the
  intent to avoid them they could not exist. They might be spurious
  but our response to them has complementary intent.}. 

\item Events that
are repeated tend to suggest greater causal intent, however, we need
to be cautious of their semantics. We do not want to conflate a
subjective attachment to certain signals with actual intent to signal.

\item Conversely, patterns that repeat often enough to dominate a scene have
little value. They represent padding, noise, spacing, etc.  Items in
the Goldilocks middle range contribute most to intent.
\end{itemize}
These ideas are reminiscent of the definitions of the Shannon entropy,
or information density \beq S = - \sum_i \; p_i \log p_i, \eeq where
$p_i$ is the probability of a process being in state $i$, as described
by Shannon's Information Theory\cite{shannon1,cover1}. However, a
probabilistic functional is insufficient to describe a dynamical
process out of equilibrium. To begin with, a probability has no sense
of time or order. It deals only overall with an average estimate of
maximal density of information per unit length of a stream. It cannot
distinguish a short episode from a long episode, not a bursty process
from a stable one.  Here, we are looking for something varying along
the stream, as a discriminator, to indicate where there is a
concentration of intentionality.

\section{Applying fractionation analysis to scene information}

A simple approach to analyzing partially coherent streams of pattern
symbols is to identify and divide the stream into discrete, elementary
alphabetic sequences. Any process may be represented as a string of
transition symbols\cite{lewis1}. Human text already has standard
alphabets of symbols; DNA has its molecular codons, etc. 
The weather has change events in pressure and temperature, and so on.
Every measurable change event is equivalent to a symbol over some alphabet.

There is a lowest `atomic' scale at which a language is merely a
string of these symbols, but language also has higher scales of
combinations of these in greater sizes and shapes.  Information and
probability analyses recognize only one scale, by default, since they
always refer to a fixed alphabet.  However, we need to involve
multiple scales to recognize the signature of intent. Networks can transmute
an input alphabet into an output alphabet (e.g. in well popularized Artificial Neural
Networks, or even in carefully constructed knowledge graphs), but starting from the geometry
of a linear sequence of events, we need to initially
separate longitudinal and transverse degrees of freedom before attending to more detailed
conceptual challenges.

With this in mind, we follow a classic analysis for representing
symbolic data\cite{lewis1}.  Let $\Sigma$ be an alphabet, whose
strings form a regular language $\Sigma^*$. We can take $\Sigma$ to be
the set of byte values from 0 to 255 for all practical machine
encodings.  From this lowest level (elementary or atomic) alphabet of
states, we can define new states by forming strings, and later by
partitioning the strings and forming new non-overlapping aggregations
(like reusable phrases or subroutines written in the lower level language), taking us from
bytes to words to strings, sentences, documents, etc, all based on a
common ground state alphabet\cite{burgesstheory}.

\begin{figure}[ht]
\begin{center}
\includegraphics[width=12cm]{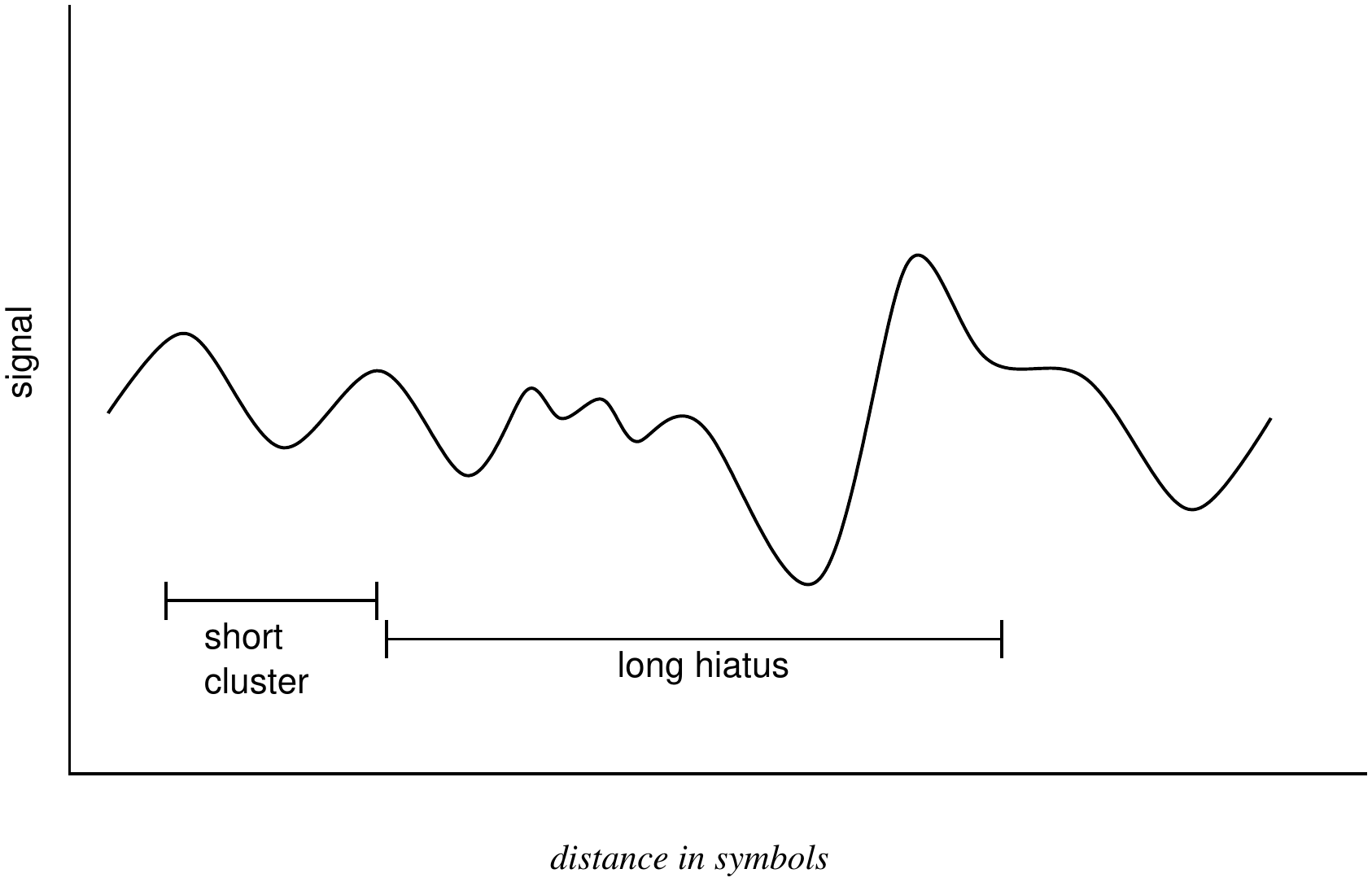}
\caption{\small The spacing between repeated signals is further
  evidence of intentionality.  Frequent repetitions all on the same
  scale tend to indicate habitual usage, padding signals (like
  spacing etc). However a hiatus followed by a repeated use over
  different timescales indicates that there is intent to repeat in a
  new context.\label{dist}}
\end{center}
\end{figure}

At each new scale or level of aggregation $\Lambda$, the
alphabet of states $\Sigma^\Lambda$ defines the set of acceptable
members, which does not necessarily include every possible combination.
Similarly, at any scale $\Lambda$, from bytes upwards, we may define
an element of the new alphabet $w^\Lambda_1$, called a 1-gram, to be a unique and
bounded configuration of the elementary alphabetic states $\Sigma$ (e.g.
pixels forming alphabetic letters, or pictograms forming phrases, codons forming genes, etc).
This leads to a derivative alphabet $\Sigma^\Lambda(\Sigma^*)$ by combinatorics. Further, we define
an $w^\Lambda_n$ to be an n-gram, i.e. an ordered sequence of $n$ 1-grams.
In natural language, 1-grams are words, and $n$-grams are text fragments or phrases of $n$ words.
We can adopt these variables at any aggregate scale. For Natural Language Processing, for example,
we have a special interest in sentence and paragraphs as the conveyances of intended meaning.

For natural language, single words or 1-grams are the elementary
particles of intended meaning, though very few single words, except
for proper names, have significance on their own.  A meaningful string
is formed only from words that are accepted as being part of the
receiver set $W^*$, i.e. they cannot be random.  Over an input stream composed
of n-grams $w_n \in W^*$, we can now define the proper time coordinate
$\tau$ of the process to be position counter of the a word within the
stream, so that over any range of $[\tau_i,\tau_f])$, called a scene
\beq
S(\tau_i,\tau_f) \in \Sigma^{\Lambda*},
\eeq
 from initial to final states,
we can count the occurrences of each n-gram $\Phi(w_n)$. 
$\Phi(w_n)$ forms a distribution of frequencies at which $w_n$ occurs
within the scene $S(\tau_i,\tau_f)$.  The length or number of words in
scene $S(\tau_i,\tau_f)$ is then \beq L = \sum_{w_1} \Phi(w_1), \eeq
so that the probability of encountering $w_1$ is $p(w_1) =
\Phi(w_1)/L$, and the Shannon base level entropy of the scene
distribution is $S_1 = - \sum_{w_1} p(w_1) \log p(w_1)$.  We can also
form the distribution of $n$-grams $S_n = - \sum_{w_n} p(w_n) \log
p(w_n)$, where $n < L$, with increasingly sparse results.

Once again, in a single scale statistical analysis, the Shannon entropy is
frequently used to express an upper limit on the information per unit
length represented in a stream. It would be not be unnatural to jump
to the conclusion that intent is a version of that conventional view
of informational entropy.  However, since entropy is measured in terms
of scale-free probabilities, which (either intentionally or
unintentionally) wipe out any knowledge of how
information varies inhomogeneously in time, one expects intentionality
to be reflected by variations in the information density for certain
symbols. If we underline a message, by repeating it
(either for clarity or for emphasis), we do not increase the amount of
information transmitted per unit length, nor its entropy, yet this is
an accepted way to signal an increase the level of intent to influence
the receiver.  This suggests that the amount of {\em effort} expended in
expressing a message is significant to its intended outcome.
This leads us back to traditional accounting methods like energetics.

\subsection{Post hoc probabilistic frequency counting}

For comparison, we begin with a classical frequency analysis.  After
the observation of a complete episode, probabilities can be computed
for data streams post hoc, and thus may only be used for prediction
after processing a scene. A frequency interpretation of probability is a flattened
summary of event occurrences, from which one can attempt a kind of
reconstructive recovery process.  

In terms of post hoc probabilities,
we can rank all fragments in a stream in terms of their global
repetition frequencies, as well as the approximate amount of work done
by the source in producing the phenomenon, e.g. the length of a
string, (or for pictogram text, the number of strokes in a character),
etc. The cost of writing or
emitting an event, a descriptive phrase, requires a measure of `work of intent' 
to overcome, like a potential barrier for an agent with limited resources.  

This ranking is what we may call the static
intentionality score for a fragment of text $w \in S(t_i;t_f)$.  If
the occurrence frequency $\Phi(w)$ has dimensions of $n$-gram per
interval of proper time (for text, this is a sentence), let us propose
that there is some threshold frequency $\Phi_0$, with the same
dimensions, representing a kind of maximal intent to shape the outcome
of a process.  There are three dimensionless cases for the frequency
of some string $w$
\begin{itemize}
\item $\Phi(w)/\Phi_0 \ll 1$: $w$ is incidental, accidental, or spurious.
\item $\Phi(w)/\Phi_0 \simeq 1$: $w$ is intended.
\item $\Phi(w)/\Phi_0 \gg 1$: $w$ is habitual padding or background noise.
\end{itemize}
A suitable approximation to the scale $\Phi_0$ may be the total number of patterns in
the scene (e.g. a low multiple of the total number of sentences in a whole
document), which is representative as an order of magnitude for
intended content. However, we find that the length of a document is not a significant
measure of intent. From statistical studies in \cite{burgess2020testingquantitativespacetimehypothesis1,burgess2020testingquantitativespacetimehypothesis2}, we find that intentionality
has a finite range, before a new idea comes along. We call this range a coherence interval.
\begin{definition}[Coherence interval $\lambda$]
The number of events over which an agent is assumed to have related intent.
\end{definition}
This interval is a property of the cognitive ability of the observer. For humans,
we find this to be close to the number we call Dunbar $D_30$\cite{dunbar2}.
Clearly there is no absolute, but an approximate average can be identified without attaching
to any particular language or writing system because we measure the stream in complete sentences,
not in word representations. This we define a coherence frequency scale $\Phi_0 = 1/D_{30}$.

When we analyze derivative strings of the language at scale $\Lambda$,
each scene of $L$ words is an ordered collection of $L/n$ $n$-gram fragments.
Our aim is to measure the intentionality associated with these fragments.
The intentionality of a string
\begin{definition}[Intentionality of a string]
The intentionality of a string is 
defined to be a weighted function of the work invested in representing the string
in terms of its $n$-grams.
\end{definition}
If the work cost invested in attempting $w$ is great, an agent may be said to intend it more
than if the work is low.
We define the work associated with an intended action as follows.
\begin{definition}[Work associated with a string]
Any intentional phenomenon
Each alphabet element in $\sigma \in \Sigma^\Lambda$ may be assigned a Work value $W(\sigma)$.
Work is a linear property of strings, so the Work of an aggregate is the sum of Work values of its elements:
\beq
W(\lbrace \sigma_1,\sigma_2,\ldots \sigma_n\rbrace) = \sum_{i=1}^n\; W(\sigma_i).
\eeq
\end{definition}
Note that, in Western alphabetic languages, all characters cost about
the same to write. Capitalized or upper case letters cost slightly more
work, so we could add a small premium factor
for the intent to emphasize these (as a type of repetition). The main work lies in the length of the word. 
In Chinese writing, and some other Asian writing systems, the number of
strokes in a character contributes to the amount of work\cite{NLnet}.

We can now formulate an expression for the intentionality of a string $w$ over some
scene interval $S(\tau_i,\tau_f)$, with these properties:
\beq
I(w,\Phi) = 
\frac{\Phi(w)\,W(w)} 
{1 + \exp{\left(\frac{\Phi(w)}{\Phi_0}-\rho\right)}}.\label{run}
\eeq
where $\rho$ is a threshold value of order $1/D_{30}$, associated with the expected density of ideas
in a scene over a coherence interval. It is conceivable that this value might vary in different kinds of
description, e.g. in natural language versus computer programs or imagery, even in different styles of text.
The numerator is proportional to the the work of intent and the occurrence frequency, magnifying 
the importance of fragments that are repeated. The sigmoid denominator curbs the increase in intentionality
by penalizing repetition over the coherence range. Since this is based on global values, it does not adapt to
regions of an episode; however, we correct this in the next section.

When adding the contributions to from a structure at scale $\Lambda$,
we could sum contributions only at a single scale. However, cognition
is a multi-scale phenomenon and our understanding of scenes depends on
the interaction of concepts born at several levels and recycled as
substituted phrases.
The $n$ parallel decompositions are analogous to processes of different orders in
a perturbative expansion.
\beq
I(s) = \sum_{n = 1}^{n_\text{max}} \sum_{i = 1}^{L(s)/n}\; I(w_i^\Lambda;\tau_i,\tau_f)
\eeq
Although an improvement over the entropy, this method shares the same
criticisms that we had over probabilistic approaches in general.
Intent has a history. It makes no sense to think of it as a Markov process.
A better approach would be to start at the beginning of a stream and follow
the stream, adapting to it rather than summarizing the whole journey
in terms of averages. We need to go from an equilibrium view to a non-equilibrium 
view of intent.

\subsection{Longitudinal burstiness in an inhomogeneous stream}

To go beyond a reconstruction from a flat distribution of frequencies,
we need to address the dynamics of cognition along the `proper time'
axis of the process.  Each time we intend something, after a hiatus,
the level of intentionality should peak, since it costs us more work
to reconceive of the idea rather than for simple copy-cat repetition.  If we repeat the idea several
times in a coherence interval, the level of intent drops because its
significance has already been counted---we wish to factor out
`symmetrical occurrence', like a factor group.  This irregular
persistence of a concept in terms of its repetition along a stream of
narrative text is a second way to measure its significance as theme of
the narrative. Probabilities cannot tell where or when a repetition
took place, only that they happened.  Intentionality implies that we
control when and how often a signal is given.  We can thus approach
intentionality using a form of longitudinal interferometry between
coherent parts of the language process.

One divides up the total text into coherence intervals, each of which
contains a `chemical admixture' of $n$-gram symbol fragments that are assumed
to be causally unrelated.
Even without a detailed knowledge of frequencies for modelling probabilities, 
we would expect padding words and even noise to be evenly distributed
along the trajectory of a stream than intentional words and phrases---which
would likely be more bursty, i.e. clustered into regions separated by
voids, because the subject of writing is not constant, it's part of a structured process.

We can detect the presence of such bursts and voids economically by
measuring the distance (in terms of sentences) between each $n$-gram
occurrence, and eliminate $n$-grams that have no significant pauses.
In other words, we are looking for a minimum difference between the
shortest distance between two cases and the longest distance between
two cases. If the largest gap is a significant fraction of the stream
or document length, by a scale factor associated with a significant
amount of information, then this offers some certainty that the phrase
is non-random, because some effort has gone into constructing the same
phrase again in a different context. The degree of effort associated with
each particle in the stream plays a role too (see next section).

This discriminator is simple, however it has a problem. For $n$-grams
longer than $n=3$, there is very little repetition. The distribution
of $n$-grams typically follows a power law in lengths so, in moderate
sized streams, finding longer clusters generally means finding only
one or two exemplars in the whole stream. This is not a statistically
significant basis on which to define probabilities. So, in the absence
of external training over potentially very different contexts, we
can't adopt a purely probabilistic approach. Fortunately, we don't
need precision to make approximate selections from small sets.  How we
rank items is less important than adopting an `intentional' policy of assigning high
intentionality to them as long rare fragments.

Any scattering of patterns, like $n$-grams in a text document forms a distribution
of locations with intra-pattern distance $\Delta(n)$. The distribution will be
`intentionally' non-uniform if it has anomalous gaps in the occurrence of 
regular terms. Again, this is unreliable to assess for small data streams.
Nevertheless, agents have no other option than to form an opinion. Statistical
frequency significance is one approach, but it has the aforementioned limitation
of being scale invariant and therefore telling us nothing about the 
dynamical stream process. A running correlation in the patterns of the significant
patterns, on the other hand, offers a second angle on which to determine whether
an occurrence is probably spurious or `random' or not strongly intended, or if it
is a recurrent concept.

\beq \Delta(n)_\text{max} > \Delta(n)_\text{min} + \langle \delta(n)
\rangle 
\eeq 

The reason this can make sense is that the scale $\langle
\delta(n) \rangle$ is not determined by a particular document. As an
order of magnitude, it is defined by the cognitive process itself.  In
order to express a certain idea, a certain amount of text is typically
needed, but this is constrained by our physiology. Using the argument
in \cite{burgessdunbarpub}, the basic scales can be based taken to
coincide with the average number of things humans can associate with
in a working context (which is one of the Dunbar
numbers\cite{dunbar2}) measured in sentences.  Again, we are not
looking for precise answers, but order of magnitude estimates based on
cognitive processes.  These can be established even without information
about the length of the input.  Unlike deep learning Large Language
Models, which seek probabilistic accuracy over large training corpuses
with even larger energy costs, we require nothing more than some basic
experience of language to make this appraisal.  The cost is negligible
and requires no training.

Machine learning techniques emphasize perfect recall, but this underestimates the
importance of forgetting. The competition between a rate of learning and rate of forgetting
determines a varying assessment of potential intent. Something repeated incessantly
is less intentional and more habitual.

A similar approach can be adopted by hardwiring a basic scale $\lambda$, for the coherence
length over which we expect a narrative text to be roughly about the same topic.
Past studies have pinned this at a few tens of sentences, sometimes rounded to 100
sentences as an order of magnitude upper limit. Here, we are motivated to draw on
the cognitive studies around learning of trust, and pick the approximate Dunbar number 
for attentive work: 45.
So,
\beq
I(w,;\tau_i,\tau_f) = W(w) (1 -e^{-\lambda(\tau -\tau_\text{last})}), ~~\tau >\tau_\text{last}.
\eeq
where $\lambda$ is the reciprocal coherence length, estimated by the working scale Dunbar number.
Apart from being simpler, immediate, and requiring less effort to compute with the same amount of memory,
this yields a result that effectively incorporates both the longitudinal and transverse criteria
for processing in a natural way. There are some differences in the selection thresholds for the
ranking of events.

\section{Intentionality as a dynamical assessment}

We can now apply these definitions to a concrete study of narrative
texts, continuing the work in
\cite{burgess2020testingquantitativespacetimehypothesis1,burgess2020testingquantitativespacetimehypothesis2}.
A practical case for intentionality is to extract the parts of a text
that are contextual for indexing and common concept identification.
These are the complementary mirror image of intentional parts.  We
thus address a very practical problem: to extract the most significant
parts of a text, based on the sense of highest intentionality. One can
imagine what that would mean for a document we write ourselves, but
what could it mean for a machine to do it for a third party?

\subsection{Symbolic fractionation method}

Fractionation is the process of splitting up a sample into small
pieces and separating them to analyze the composition---as in organic chemistry, or DNA analysis.  A narrative
text is a coherent stream of text that we can split up into $n$
parallel streams, formed from fragments of $n$-grams. These behave as
independent decompositions, but with some similarities.  Each type of
fraction satisfies a power law distribution over $n$\cite{burgess2020testingquantitativespacetimehypothesis1}, 
and thus there
are exponentially fewer $n$-grams with higher $n$. By the time we
reach $n=4$ there is little repetition, thus a high degree of
uniqueness and intentionality. Longer $n$-grams behave like proper names
and are thus likely places to find the beginnings of invariant concepts.
Low $n$ fragments form an ambient soup around
the longer fragments that support their meaning, but can be omitted initially to
distill core building blocks on a `molecular' level.  

Separating these as partially independent fragments, and excluding
some words that are the highest frequency spacers, we can use the
$n$-grams ($n=1,\ldots,5$) as a spanning set for labelling the
representation of concepts or themes within the text. They define the
dimensions of a primitive feature space.  The scale of data is central
to this approach.  If there are too little data, then everything is
related to everything else, or everything is anomalous---this is why
we work in measures of coherence intervals. Human cognition has famous
limitations that have been used to explain human trust
behaviour\cite{burgessdunbarpub}, including the Dunbar numbers for
cognitive attention and the short term memory limit for indirection at
a handful of levels. Artificial systems do not have such limitations,
but if we want them to work for humans we need to respect our own
limits.

Although all fragments in a stream are intentional to some extent, we
can distinguish thematic fragments that are contextual, i.e.
correlated by exogenous circumstances, from those that are unique,
original, and anomalous in the context.  Context is intuitively a cloud of
secondary signals around a core concept, offering a blunter search
region, but possibly offering a broader way in to find intended
statements which are the target of special interest. This is often
where key proto-concepts are to be found.  In other words,
ambient context forms the bases for random access memory keys by which an agent
might attempt to recall the details later.  Intentional fragments, on the other hand, are
all those which are not contextual by frequent repetition, so they follow on from
basic entry points by causal reasoning. 

In the Semantic Spacetime model\cite{burgess_sst12025}, one can view this as
defining an agent in terms of its boundary between interior and
exterior, we can effectively partition information about the agent
into two approximate parts: context and intentional. This is what Searle refers to as
extensional and intensional information\cite{searle1}.
A common strategy in statistical mechanics is to separate dynamics into
fast and slow variables. Rapid fluctuating changes are often associated with
noise to be averaged away, while longer term trends are treated as persistent information.
In a similar way, the separation of independent agents and concepts is made possible
either by the sparseness of their embeddings in space and time, or by a significant
boundary layer such as a cell wall. The separation of semantic content into intentional and contextual
variables is a similar strategy.

\subsection{Narrative text test data}

Applying the
method to a number of freely available texts, and some of my own
texts.  We can separate the identifiable $n$-grams into two classes:
those with a high likely hood of being unique and important and those
repeated more often as supporting ideas.  We should also be clear
that, intentionality is not a binary assessment.  We can call the two
parts `intentional' and `contextual' but that does not mean that
context is unintended.  Rather, it means a reflection of a more common
ambient low level of environmental intent, not a complete absence.
One imagines assessing the intent of everything from briefly scribbled
notes about a scene to carefully worded narrative texts. The following
source texts were used for this demonstration:

\begin{center}
\begin{tabular}{|c|c|c|l|}
\hline
Characters & Words & Name\\
\hline
1206540 &  208458 & Moby Dick\\
1150632 & 192106  & The Origin of Species (6th)\\
1050785 &  179511 & History of Bede\\
31301   & 5193 & Thinking In Promises\\
26385   & 4601  & Obama inaugural speech\\
2046    & 297     & A page from a diary\\
\hline
\end{tabular}
\end{center}

\begin{figure}[ht]
\begin{center}
\includegraphics[width=7cm]{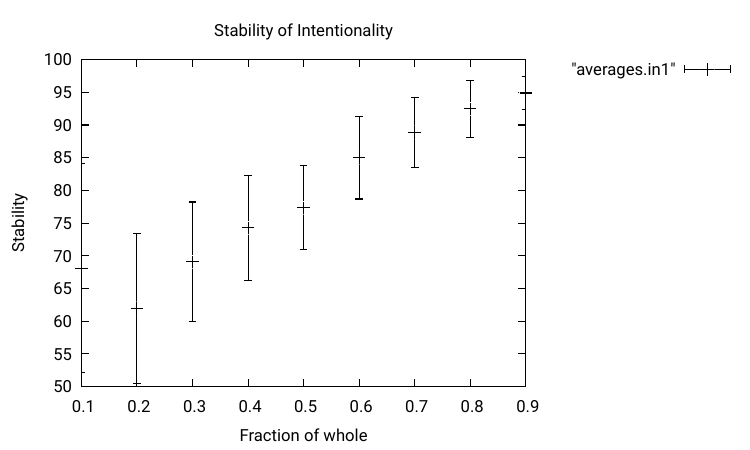}
\includegraphics[width=7cm]{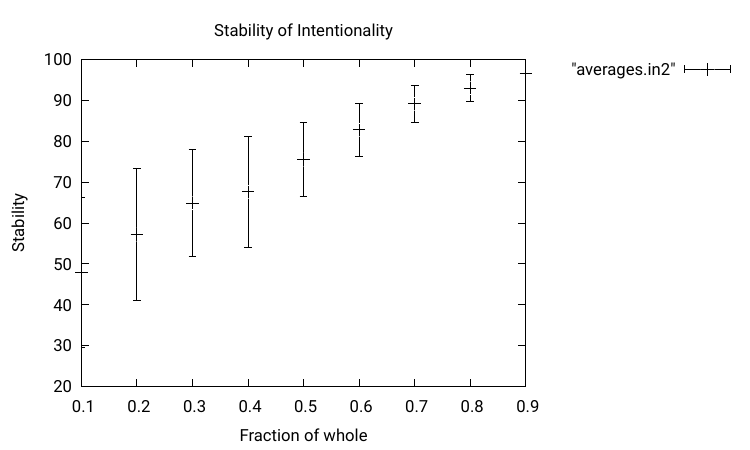}
\includegraphics[width=7cm]{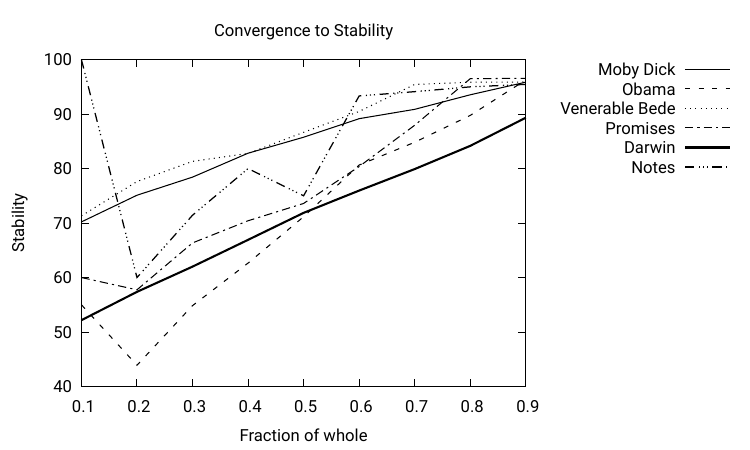}
\includegraphics[width=7cm]{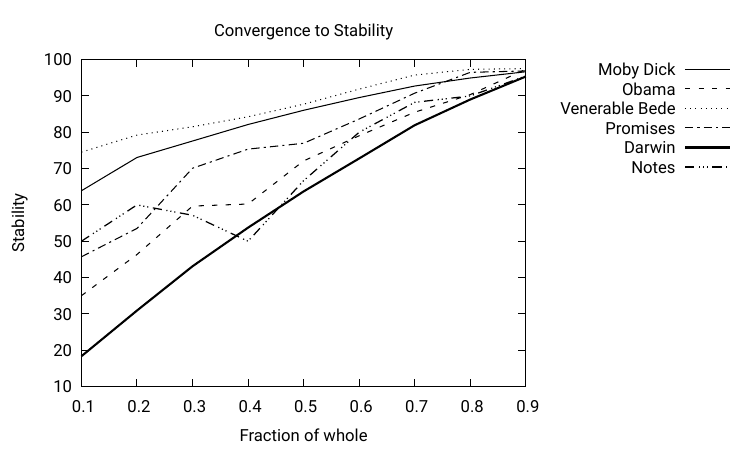}
\caption{\small Plot giving a heuristic impression of the degree of overlap
in fragments selected by $I(w,;\tau_i,\tau_f)$ and $I(w,;\Phi)$ for the documents studied.
Upper panels show average and deviation of diverse document lengths and types,
giving a rough idea of the spread of intentionality using different models.
The left panels are calculated using the expression in (\ref{run}); 
on the right are the counterproposal for intentionality decaying without resetting
with distance, which has the effect of relying on first impressions.
The left hand average results begin on average around the 50s, while the right hand side begin in the 20s,
but as we digest more and more of the stream the reliance on intent converges quickly
to a similar result.
The shortest documents show the most precarious variability, 
since they have less to build an assessment on, while the two long texts
(Moby Dick and Darwin's Origin of Species) give simple linear behaviour as
one might expect.\label{cmp}}
\end{center}
\end{figure}

The intentionality functions can be examined over these sets
to show their behaviours. Given the expected high noise levels in
text, the remarkable linearity of the calibration over long texts (figure
\ref{cmp}) is a result of the linear partitioning into coherence
regions. This is partly a choice, however because the length of a
sentence is not fixed, it is also a reflection of the cognitive
stability of human attention, reflected in the Dunbar numbers.

There are four distinct approaches to compare in order to gauge the level
of intentionality in a symbolic process:
\begin{itemize}
\item Ranking $n$-grams by global (Markov) frequencies over a complete stream.
\item Ranking $n$-grams by running (memory) frequencies over a complete stream.
\item Using global coherence and spectral content of $n$-grams to distinguish change globally.
\item Using running coherence and spectral gradient of $n$-grams to distinguish change locally.
\end{itemize}
In the first pair, we use a numerical score to rank the importance of fragments of different sizes post hoc
and in realtime,
on the understanding that the longer the phrases (the greater $n$) the more unique
they are. In second pair, we use primarily longitudinal coherence as the distinguishing
dynamic for ranking importance, both post hoc and in realtime. An intentionality score based
on work and frequency can be used in a secondary way to rank the surviving $n$-gram fragments.

A second level of intentionality can be found by coarse graining the process of coherence intervals.
Apart from being consistent with the model that ranks intent potential, this allows us to form
a running split between ambient meaning and focused decision making, or `thinking fast and slow'
to paraphrase Kahneman\cite{kahneman}.

\begin{figure}[ht]
\begin{center}
\includegraphics[width=10cm]{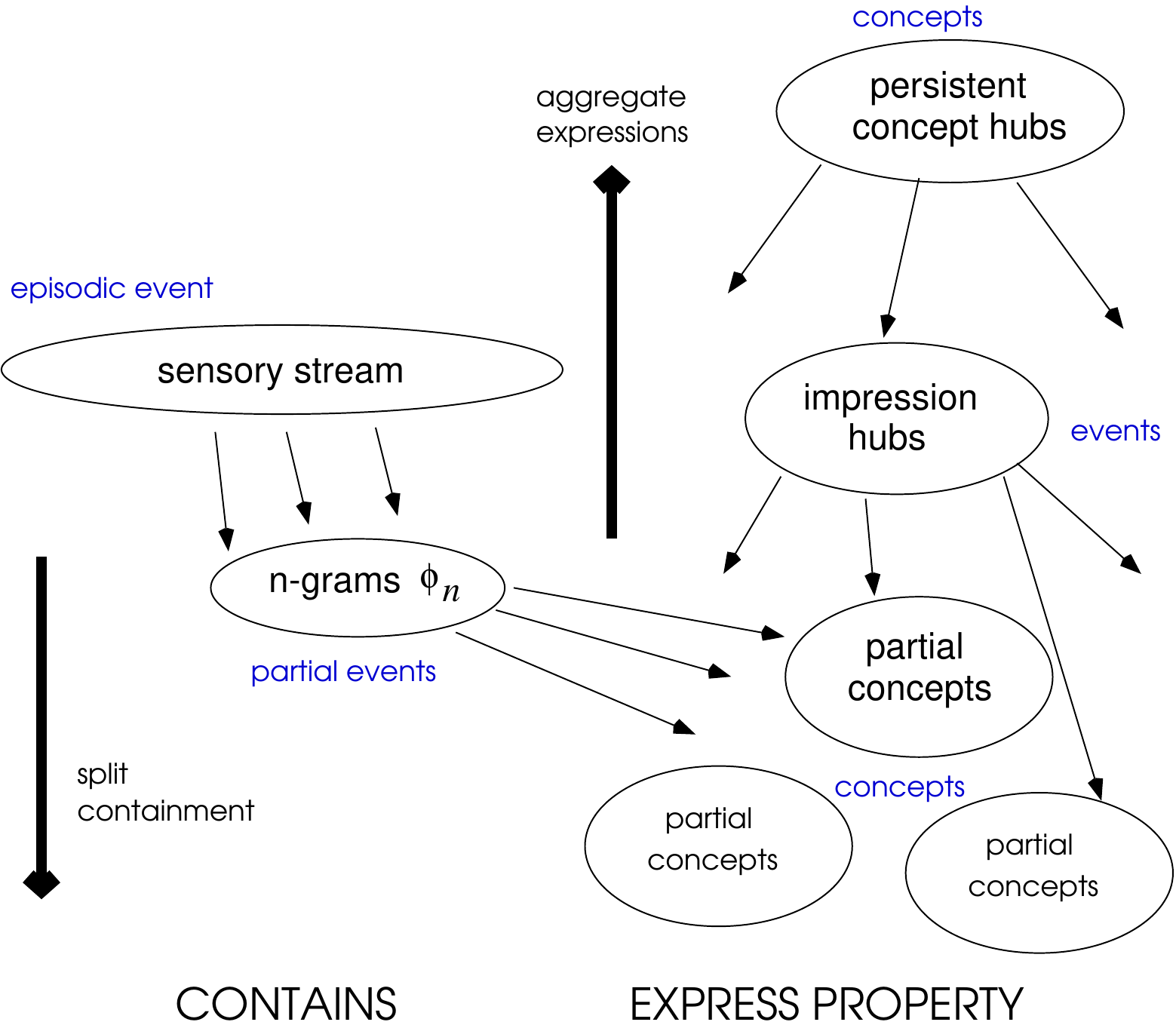}
\caption{\small The process of transmuting episodic sensory signals info fractions, then
low level impressions associated with them, then collective impressions
about a coherent interval of time, and finally distilling these into long term
persistent concepts over a much longer time frame.\label{concept}}
\end{center}
\end{figure}

Inspecting the output of fragments with a human eye, someone's one method
appears appealing, sometimes the other.
The longitudinal coherence method identifies the split between
contextual and intentional, from the gaps in between repetitions, but it doesn't
order them. Those fragments that can then be 
post-ranked according to the static intentionality to present them in some order.
If we are in doubt, a change is counted as intentional, so the cognitive work is not
so much about ranking intent as in discovering what to ignore in the ambient field.

Each document behaves rather differently in terms of ambient context
versus anomalous or `particularly intended' fragments. This is to be
expected.  Style plays a role. The text by Obama is rich in
intentionality, with very little ambient context, as one might expect
from a prepared directed speech.  Context also requires a longer
document to build up a sense of repetition.  One learns quickly that
there are no universal meanings or intrinsic values for intent.
There is no universal agent. The best we can do is to rank agent's
effort against itself.

In section \ref{episode}, we look at the interval by interval selection of fragments that
could be used to index the context of paragraphs in a book, or define the
cognitive context of a scene, `frame by frame'. However, skipping to the end first, 
the summary of the documents after a full reading is interesting.
This is like reviewing what we learned from reading it. Here are of the top samples
for ambient context and anomalous intent from the texts. Can we tell which
text they might have come from?

\begin{center}
\begin{quote}\em
{$n=1$ \sc ambient}: 
classification,
domestication,
modifications,
organisation,
circumstances,
nevertheless,
archipelago,
variability,
development,
constitution,
differences,
productions,
inheritance,
inhabitants,
individuals,
naturalists,
descendants,
parent-species,
consequently,
distribution,
crustaceans,
competition,
extinction,
difficulty
\end{quote}
\end{center}

\begin{center}
\begin{quote}\em
$n=1$ {\sc anomalous}: 
reproduction,
difference,
arrangement,
progenitors,
intercrossing,
pedicellariae,
considerations,
changes,
organisms,
resemblances,
numerous,
perfect,
consideration,
relations,
concerned,
branches,
cattle,
expected,
favourable,
finally,
anticipated,
butterflies,
improvement,
perfected,
district,
observed,
peculiar,
preservation,
difficulties,
perhaps,
giraffe,
itself,
elevation,
principles,
related,
closely,
investigation,
contingencies,
germination,
competitors,
experiments,
naturalised,
widely,
causes,
representatives,
palaeontologist,
observations,
\end{quote}
\end{center}

\begin{center}
\begin{quote}\em
$n=2$ {\sc anomalous}:
perfectly fertile,
north america,
intermediate gradations,
domestic productions,
surrounding conditions,
geological formations,
favourable variations,
physiological importance,
dominant species,
other plants,
same country
\end{quote}
\end{center}

\begin{center}
\begin{quote}\em
$n=3$ {\sc ambient}:
through natural selection,
conditions of life,
throughout the world,
from each other,
struggle for existence,
state of nature,
divergence of character,
descent with modification,
habits of life,
closely allied species,
intervals of time,
varieties and species,
forms of life,
animals and plants
\end{quote}
\end{center}

\begin{center}
\begin{quote}\em
$n=3$ {\sc anomalous}:
oscillations of level,
principle of inheritance,
number of generations,
our domestic productions,
modified and improved,
intermediate in character,
with each other,
our domesticated productions,
means of transport,
have been modified,
\end{quote}
\end{center}

It is not hard to see that these fragments match Darwin's study of evolution.
In the inevitable Moby Dick, we find fragments that include:

\begin{center}
\begin{quote}\em
$n=2$ {\sc ambient}: 
sperm whale,
white whale,
captain ahab,
moby dick,
cried ahab,
his head,
cried starbuck,
with him,
before him,
captain peleg,
cried stubb,
right whale,
old man
\end{quote}
\end{center}

\begin{center}
\begin{quote}\em
$n=2$ {\sc anomalous}: 
whale fishery,
whaling voyage,
his hand,
about him,
aloft there,
for example,
first place,
this matter,
among them,
start her,
his cabin,
over him,
this whale,
once more,
his pivot-hole,
his forehead, ...
\end{quote}
\end{center}
We see that many intentionality values lie close together and are basically
indistinguishable.
In a Promise Theory book by the author, the final ranking includes:
common concepts in Promise Theory
\begin{center}
\begin{quote}\em
$n=2$ {\sc ambient}: 
promise theory,
incomplete information,
intended outcome,
own behaviour,
other words,
for humans,
this book, ...
\end{quote}
\end{center}

\begin{center}
\begin{quote}\em
$n=2$ {\sc anomalous}: 
distributed information,
hourly checks,
absolute certainty,
already happened,
starting point,
atomic theory,
don't control, ...
\end{quote}
\end{center}

In each of these cases, the goal is to extract a list of fragments
that allow us to `index' or secure an entry point, on random access,
to a longer narrative based on impressions of it.  This could be
applied to any stream, but here I only use text streams as an
application for the SSTorytime project\cite{sstorytime}.
If we reach for a high level understanding of intent, we might
discount the importance of these low level fragments and seek a
linguistic paraphrasing like ``Captain Ahab wants to kill the white
whale''. However, no such sentence exists in the text, so it would have to be
invented. This is what the Large Language Models can do, however
this happens at the cost of enormous processing and training.
A more important question, in the understanding of intentionality, is
how would one represent that concept without advanced linguistics---just
from the fragments we've produced?

\begin{figure}[ht]
\begin{center}
\includegraphics[width=10cm]{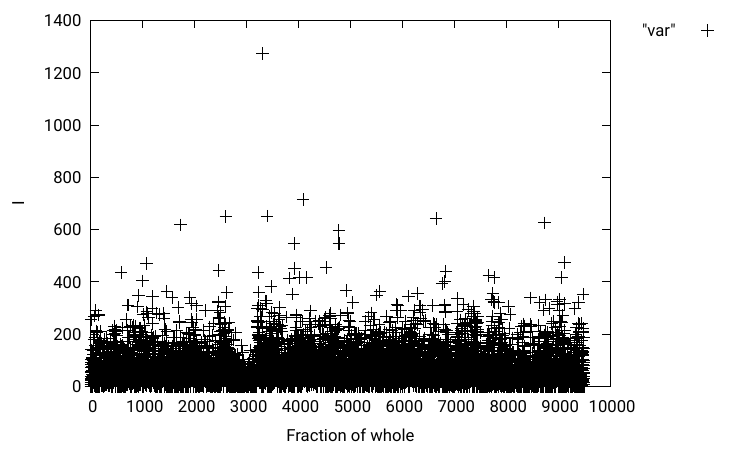}
\caption{\small Plot of running $I(w,;\tau_i,\tau_f)$ over a long document, showing that
there is no obvious pattern to the level of intentionality in consecutive sentences. 
Intentionality appears as a random variable, because natural language is
highly dense in associative meaning. Sampling
the document in summary focuses on skimming the most significant sentences,
hoping that process energetics can serve as a proxy for inference of meaning.\label{intent}}
\end{center}
\end{figure}

\section{Unfolding of knowledge and concepts along an episode}\label{episode}

By looking at natural language (only English here, but representative
for many alphabetic text with space separation) we have a simple way
of gaining insight by direct recognition\footnote{There is a subtlety
  for languages that spread phrases, e.g. by sending part of a verb to
  the end of a sentence in ways that can affect the whole meaning. Then agents
are forced to remember more to comprehend meaning.}.
Each reader becomes a potential receiver and can gauge their own
experience.  In the real world we experience (even in professional
contexts like research) how different levels of intent result in
different interpretations.

The tension between intentional and habitual (contextual) affects our
actions in complex ways.  Perhaps a majority of actors will tend to
recycle well-known material by adding to a corpus of common knowledge
in context. Only a minority may invest the greater cognitive effort to
intentionally reinterpret and come up with new concepts.  Thus, along
any semantic information channel the intentionality potential plays a
role, importantly, for both sender/promiser (+) and receiver/promisee
(-).

Acquiring knowledge typically requires an intention to learn.
Repeating, revisiting, and rehearsing of knowledge is what allows us
to `know something as we know a friend'.  Why?  From written or spoken
language, we have to transmute actualized versions of some
representation into conceptual versions. The four vector types of
Semantic Spacetime (follows from, contains, expresses property)
represent different stage of this process. Initially, a sensory string
is an exterior physical actualization of some process (sounds, writing, etc), but we need to transmute that
into an interior conceptualization (see figure \ref{concept}).
\begin{itemize}
\item The actualized form of a statement is an event, which {\sc contains} fragments, representing include proper names, phrases,etc. Fragmenting releases contained composition.
\item We split into fragments as events contained by the full string.
\item The events {\sc express} different meanings, singly and in combination. This is a learned association.
\item These meanings are concepts, which can be aggregated into hubs, in different ways, 
as larger concepts which {\sc express} them. Recombination of ideas expressed forms a model.
\item The common phrases signal their meanings both generally and specifically, which tends to distort
and generalize the meanings we perceive in the transmutation, because the mapping is not one to one.
Signals are multivalued.
\item The receiver has to select a meaning from the possibilities based on its current context state.
\end{itemize}
Thus the full deconstruction of received signalling or communication is a multistage process
according to the Semantic Spacetime hypothesis.

In the earlier study\cite{burgess2020testingquantitativespacetimehypothesis1,burgess2020testingquantitativespacetimehypothesis2}, I made the observation that---after condensing the texts progressively---the simple 
and direct linguistic meanings of the strings tended to give way to more emotional categorizations.
A novel, like Moby Dick, in the end was not about whales but about human feelings of anger and vengeance.
Those themes emerged from the distillation of small fragments. Similarly. Darwin's text was about awe, Bede's account was about religious feelings, and so on. This seems to occur because the fragmentation into basic
1-grams still leaves an impression, which is more ambiguous and expansive.
Try gauging meaning from 1-grams alone, and the result is something more instinctive.

Intentionality plays a role in the scale at which we understand the
signals because the intent is both on the side of the transmitting
source and on the side of the receiver (promiser and promisee, in
Promise Theory language).  Experience over many episodes clearly
contributes the most to the development of understanding. Foreigners
can learn a new language and still understand nothing of what's said
because there is a cultural level to communication too in which
apparently meaningful phrases as co-opted as proper names for very
different concepts (`break a leg', etc).  
We can observe this in action.

Consider the very opening passage of Darwin's book---just the very
first coherence interval of 45 sentences.  These phrases of different
order have lead with the lowest reader intentionality and require
progressively more effort.  In we skim a scene, with low intent to
analyze is repeatedly, we capture mostly the first order fragments and
their immediate associations as a bag of words. We lean on the common
meanings as proper names.  As readers, we know more than a machine or
lower organism would about the meanings of these patterns, so we can
try to imagine what different readers might experience, from a
non-English speaker to a young or old reader, less or more educated,
etc.
\begin{enumerate}
\item We begin with the lowest level of receiver intent to divine sender intent:
the ambient common words we first encounter are:
\begin{center}
\begin{quote}\em
{\sc ambient}: domestication, important, variations, climate, variation, subject, 
food, birds, structure, however, conditions, namely, world, animals,
nature, varieties, species, 
\end{quote}
\end{center}
From this collection, we might conjure an image of a bird/animal-keeper with
some scientific interest.

\item Next we go a step further, intentionally investing in the
less familiar phrases. 
As experienced meta-observers, we recognize that words like ``nevertheless'' are emphatic signalling 
the intent of the author to make a point.
\begin{center}
\begin{quote}
\em
{\sc anomalous}: nevertheless, archipelago, inheritance, naturalists,
distribution, consequently, conclusion, existence,
therefore, continent, illustration, probable, naturalist,
increase, chapter, possible, intercrossed, considerations,
modified, relations, instinct, character, created, selected,
record, trees, being, alone, habit, beings, given, itself,
organ, imperfect ...
\end{quote}
\end{center}
From this next level we immediate get a more specific sense,
referring to archipelago islands which suggest far away places.
Scientific and analytical terms already about, so we have the impression
of some kind of thoughtful analysis.

We might be surprised that these less readable words are
actually more intentional than their previous set, since they are
less immediately recognizable to a common reader,
however this is the
counter intuitive point: intent places on the leading edge of
exploration, not in the realm of comfortable normalcy.
This small step immediately recognizes `naturalist'
and a number of more scientific words, so we a taken by a feeling
of some science going on.

\item 
If we invest further intent in going down the fractionating column to $n=2$ fragments,
we obtain only two phrases in the first coherence interval:
\begin{center}
\begin{quote}
{\sc ambient}: \em under domestication, more important
\end{quote}
\end{center}
These ambient phrases don't say much, except for the concept of domestication, 
which again suggests an animal-keeper.

\item Investing further in our intent to comprehend, we immediate step up a level
of meaning, with clues about places and further evidence of a scientific mentality.
The samples here are not complete, only the highest ranking terms. We know that
intentional effort is proportional to the amount we parse too. Skimming versus
questioning is the personal intentional choice of the reader.
\begin{center}
\begin{quote}
{\sc anomalous}: \em organic beings,
south america,
this subject,
geographical distribution,
for existence,
geological record,
other species,
independently created,
for instance,
external conditions,
been modified,
slight variations,
malay archipelago,
when intercrossed,
hereafter see,
this volume,
vegetable kingdoms,
fourth chapter,
possibly survive,
geological succession,
least possible,
modified form,
innumerable species,
succeeding chapters
\end{quote}
\end{center}
Some fragments written with intent (e.g. hereafter see) may not be not strongly habitual
to the author (hence they appear in this classification) but they are not proper names
so could be dismissed by a casual reader, but be seen as evidence of
a focus on succession and change to a more probing reader. 

\item By the time we read $n=3$ there are few ambient phrases as the chance of encountering
a precise combination is exponentially small.
\begin{center}
\begin{quote}
state of nature, variation under domestication
\end{quote}
\end{center}
We now have major clues from these contextual common phrases (only two!). 
The opening and closing of a strongly intentional text
(in this case a kind of thesis) are typically where intentionality is strong in
both writer and reader. Here we already see evidence of natural selection in the hands
of intended and unintended curators!

\item Specifically more intentional phrases:
\begin{center}
\begin{quote}
{\sc anomalous}: \em conditions of life,
throughout the world,
struggle for existence,
divergence of character,
forms of life,
been independently created,
principle of inheritance,
shall hereafter see,
have been modified,
successive slight variations,
can possibly survive,
origin of species,
chance of surviving,
varieties when intercrossed,
laws of variation,
from these considerations,
distinct organic beings,
favourable to variation,
may be true,
other such facts,
their geographical distribution,
modification and coadaptation,
elaborately constructed orga
\end{quote}
\end{center}

\item By the time we reach $n=4$, there are no ambient phrases left. The degree
of intentionality has crossed a threshold and it imposes a greater
burden on the reader to comprehend the specific fragments.
\begin{center}
\begin{quote}
{\sc anomalous}: \em animal and vegetable kingdoms,
than can possibly survive,
inhabitants of that continent,
strong principle of inheritance,
which i have arrived,
affinities of organic beings,
beings throughout the world,
better chance of surviving,
improved forms of life,
selection successive slight variations,
species--that mystery of mysteries,
commonly neglected by naturalists,
recurring struggle for existence,
several distinct organic beings,
beings inhabiting south america,
not been independently created,
effects of external conditions,
called divergence of character,
nourishment from certain trees,
conclusions have been grounded,
conceivable that a naturalist,
must necessarily be imperfect,
justly excites our admiration,
transported by certain birds,
extracts from my manuscripts,
most favourable to variation,
refer to external conditions,
manner profitable to itself,
possible cause of variation ...
\end{quote}
\end{center}
\end{enumerate}

This small sample gives some idea of the process investment required to
comprehend the text stream, even on a basic level. As readers of English
we are cheating when we know the meanings from our `out of band'
training. A machine starting from nothing cannot know that
but a reading level human can. This, of course, is why training is required
to approach our human level of comprehension. Yet, by jumping too quickly
to a reliance on learned semantics, we miss the opportunity to view comprehension
from the purer perspective of the dynamical cognitive process, which is common
to all organisms on some level.
We are forming a map. What
questions can we ask the map?

\section{Conclusions}

A process-descriptive view of intentionality in sensory streams, with
only a Tiny Language Model based on multi-scale process rates
remarkably well to a human eye. Without actually understanding
anything, it is able to discriminate interesting agent signals and
infer proto-concepts from fragments in a data stream.  The approach is
qualitatively similar to recent arguments successfully used to define
and quantify trust in social
processes\cite{trustnotes,burgessdunbarpub}.  Intent may be thought of
as a `vector' of fragments, posited to represent a coherent assessment
of signal information; it is dimensionally analogous to a momentum in
mechanics.  Intentionality, on the other hand, is the corresponding
density with which signal contains this reactance; it is undirected
and analogous to an `energy' or rate of `work expenditure'
superficially inferred for the agent originating the information.

Traces of behavioural intent persist across multiple scales of a
cognitive signal. This mirrors the residual way that an animal reflects both
the genetics and upbringing of its parents, or a city retains the
intentions of its planners as well as its inhabitants. The retention
of intent is not a high fidelity encoding of unambiguous data,
but an approximate alignment of concepts, working statistically
against the forces of noise. The Downstream Principle of Promise
Theory tells us that the agent assessing the intentionality is itself
party to the assessment by its own intent to work.

The memory capacity of an agent is an important capability used to
assess intentionality. This is typically thought of as the resources
an agent has internally (e.g. its brain). However, most of the memory
available to organisms is external (stigmergic) in the configuration
of its environment, e.g.  writing, pheromone trails, configurations of
resources, etc.  Dunbar has shown that memory and social involvement
are correlated in social groups where both trust and intentionality
are at work\cite{dunbar1}.  Language is a prime example of something
perpetuated outside of humans. Without societal memory, and the
extended processing on that scale, languages fail to evolve and die
out altogether. Cognition is never independent of its ambient setting.
With this promise theoretic view, it is not wrong to anthropomorphize
`inanimate' processes.  An observer might feel a storm's behaviour's
targeted (i.e.  anthropomorphized as deliberate harm), but this is not
actually the observer's opinion about the possible degree of `agency'
of the weather in singling out a victim, but an assessment of the
changing impact on the observer, equivalent to a process that could
have been the result of malice.  Conversely, a lazy summer afternoon
has low intentionality, as nothing distinguishes one moment from the
next.

We should not go too far in asserting the universal importance of this
highly superficial assessment relative intentionality, because it can
never be based on first hand information. We can't infer the
abilities of an observer either.  For example, it is not easy to
estimate the actual effort exerted by an author in the writing of a
book. The author could have been dyslexic, or be writing in a second
language, etc. The effort only has significance in a relative way,
within each independent stream, and the intentionality we define is a
somewhat ad hoc assessment, not a rigorous computation.  More
importantly, the ability to extract the `chemistry' of the intentional
fragments has direct practical applications in knowledge
representation.

The current fascination with `AI chatbots' is an interesting case
where we see intentionality in randomly generated strings.  Our desire
to follow without intent feels momentarily intentional, as with any
new relationship, but later it dwindles into indolence: after the
initial intent to explore a novel relationship (even with a
contraption) we might lean too heavily on it for assistance, which
encourages a further loss of intentional behaviour.  Intentionality is 
our sense that we need to work for our own gain. In pursuit of
human self-improvement, it's only when we have the intent to learn
something that we truly remember it in its proper context. It costs us
effort, but that's what pays off.

\bigskip

A software implementation of this work is available at \cite{sstorytime}.

\bibliographystyle{hplain}
\bibliography{spacetime,bib}

\end{document}